\documentclass[lettersize,journal]{IEEEtran}
\usepackage{amsmath,amsfonts}
\usepackage{algorithmic}
\usepackage{algorithm}
\usepackage{array}
\usepackage[caption=false,font=normalsize,labelfont=sf,textfont=sf]{subfig}
\usepackage{textcomp}
\usepackage{stfloats}
\usepackage{url}
\usepackage{verbatim}
\usepackage{graphicx}
\usepackage{cite}
\usepackage{booktabs}
\usepackage{subfloat}
\usepackage[T1]{fontenc}
\begin{document}

\title{AdvMask: A Sparse Adversarial Attack Based Data Augmentation Method for Image Classification}

\author{Suorong~Yang, 
			Jinqiao~Li,
			Jian~Zhao,
			and Furao~Shen*\thanks{* Corresponding author.}
%\thanks{This work was supported in part by the National Key R\&D Program of China under Grant 2021ZD0201300, and by the National Science Foundation of China under Grant 61876076.}
\thanks{Suorong~Yang, Jinqiao~Li, and Furao~Shen are with State Key Laboratory for Novel Software Technology, Department of Computer Science and Technology, Nanjing University, Nanjing 210023, China (e-mail: sryang@smail.nju.edu.cn; lijq@smail.nju.edu.cn; frshen@nju.edu.cn)}
\thanks{Jian Zhao is with the State Key Laboratory for Novel Software Technology, and School of Electronic Science and Engineering, Nanjing University, Nanjing 210023, China (e-mail: jianzhao@nju.edu.cn)}
}

\maketitle

\begin{abstract}
Data augmentation is a widely used technique for enhancing the generalization ability of convolutional neural networks (CNNs) in image classification tasks.
Occlusion is a critical factor that affects on the generalization ability of image classification models.
In order to generate new samples, existing data augmentation methods based on information deletion simulate occluded samples by randomly removing some areas in the images.
However, those methods cannot delete areas of the images according to their structural features of the images.
To solve those problems, we propose a novel data augmentation method, AdvMask, for image classification tasks.
Instead of randomly removing areas in the images, AdvMask obtains the key points that have the greatest influence on the classification results via an end-to-end sparse adversarial attack module. 
Therefore, we can find the most sensitive points of the classification results without considering the diversity of various image appearance and shapes of the object of interest.
In addition, a data augmentation module is employed to generate structured masks based on the key points, thus forcing the CNN classification models to seek other relevant content when the most discriminative content is hidden.
AdvMask can effectively improve the performance of classification models in the testing process.
The experimental results on various datasets and CNN models verify that the proposed method outperforms other previous data augmentation methods in image classification tasks.

%We demonstrate consistent classification accuracy improvements on various CNN models and various open datasets.
%The proposed method outperforms other information deletion based data augmentation methods on image classification tasks.

% seek other relevant content when the most discriminative content is hidden
%we propose using sparse adversarial attack to locate the key points because the the calssification results are sensitive to these adversarial attack points. 
%Some invisible perturbations on these points can cause classification models to misclassify.
%However, existing information deletion based data augmentation methods often randomly deletes information in the image and thus is not targeted or interpretable.

\end{abstract}


\section{Introduction}
Image classification~\cite{image_class, image_class2} aims to classify digital images automatically, which has been a core topic of computer vision for decades.
Image classification can be applied to various fields, such as object detection, video classification and segmentation.
Recently, convolutional neural network (CNN) has made tremendous progress in image classification.
In order to further improve the classification accuracy, more complex deep neural networks, such as ResNet-152~\cite{resnet}, have been proposed.
However, if a model is too complex, for example, when the complexity of neural network is higher than that of training data, over-fitting will damage the classification accuracy.
At the same time, images in such tasks often undergo intensive changes in appearance and varying degrees of occlusion,  which are the critical influencing factors on the generalization ability of CNNs~\cite{randomerasing}.
As most image datasets only have clearly visible images, the generalization ability of classification models degrades significantly in this case.

Considering the extreme difficulty of data collection in real-world tasks, data augmentation~\cite{keepaugment,autoaugment,augment,augment2,randaugment} has been proposed as a very important technique to improve the generalization ability of CNNs by generating more effective data based on the existing data.
 
Recently, data augmentation methods based on information deletion, such as Cutout~\cite{cutout}, Hide-and-Seek (HaS)~\cite{has}, and Random Erasing~\cite{randomerasing}, have been widely used.
It is known that this kind of data augmentation methods can reduce models' sensitivity and strengthen the generalization ability of the models.
Especially, they can predict occluded objects when the original training data are all clearly visible.
The core idea of the methods based on information deletion is to generate new data by deleting some areas of the image, which can simulate the situation that objects are partially covered and the whole structure is incomplete in the real world.
However, if is found that, instead of customizing the mask according to the structural characteristics of the images, the existing data augmentation methods based on information deletion can block some random areas in the images. 
Although this may improve the classification accuracy because the diversity of training data increases, this kind of mask generation is not targeted and uninterpretable. 
In the bad situation, the occluded areas are all in the foreground, thus the objects of interest in the image are completely occluded, which seriously damages the generalization ability of CNN models. 
In addition, different images have different structures. 
In other words, the key information used by CNN models to classify different images is different.
It is much more beneficial to customize the mask for each image according to the structure information.
However, random information deletion cannot solve such problems.

Based on the above discussions, we propose a data augmentation method (AdvMask) based on information deletion, which can regularize CNN training and improve the performance of models in image classification.
The most important motivation for AdvMask is to simulate the situations where the key information for classification is partly or completely lost.
The key information in the image is closely related to the classification results, and the change of those information will lead to misclassification.
So we need to locate the points where perturbation would make the classification result different from the truth.
Adversarial attack is to deceive the target model by adding adversarial perturbations to the original clean images.
In addition, sparse adversarial attack can only perturb the pixels that have the greatest influence on the classification results.
Therefore, in AdvMask, we design a sparse adversarial attack module to find the pixels that have the greatest influence on the classification results.
For data augmentation, we generate augmented samples by occluding some of these adversarial attack points in the images.
There are three advantages of occlusion on these adversarial attack points.
Firstly, the deleted areas generated by AdvMask are based on the adversarial sensitive region, so the  deleted areas are not limited to a continuous area, but multiple areas with various shapes.
Secondly, the deleted areas generated by AdvMask is customized for each image, rather than randomly generated, so the whole deleted areas are closely related to the image structure, including both foreground and background.
Finally, AdvMask uses sparse adversarial attack to find the most sensitive areas in the image, instead of only focusing on the area where the main object is located.
At the same time, we also avoid excessive deletion and reservation of continuous regions, in order to balance object occlusion and information retention~\cite{gridmask}.
\begin{figure*}[ht]
	\centering
	\includegraphics[width=0.9\textwidth]{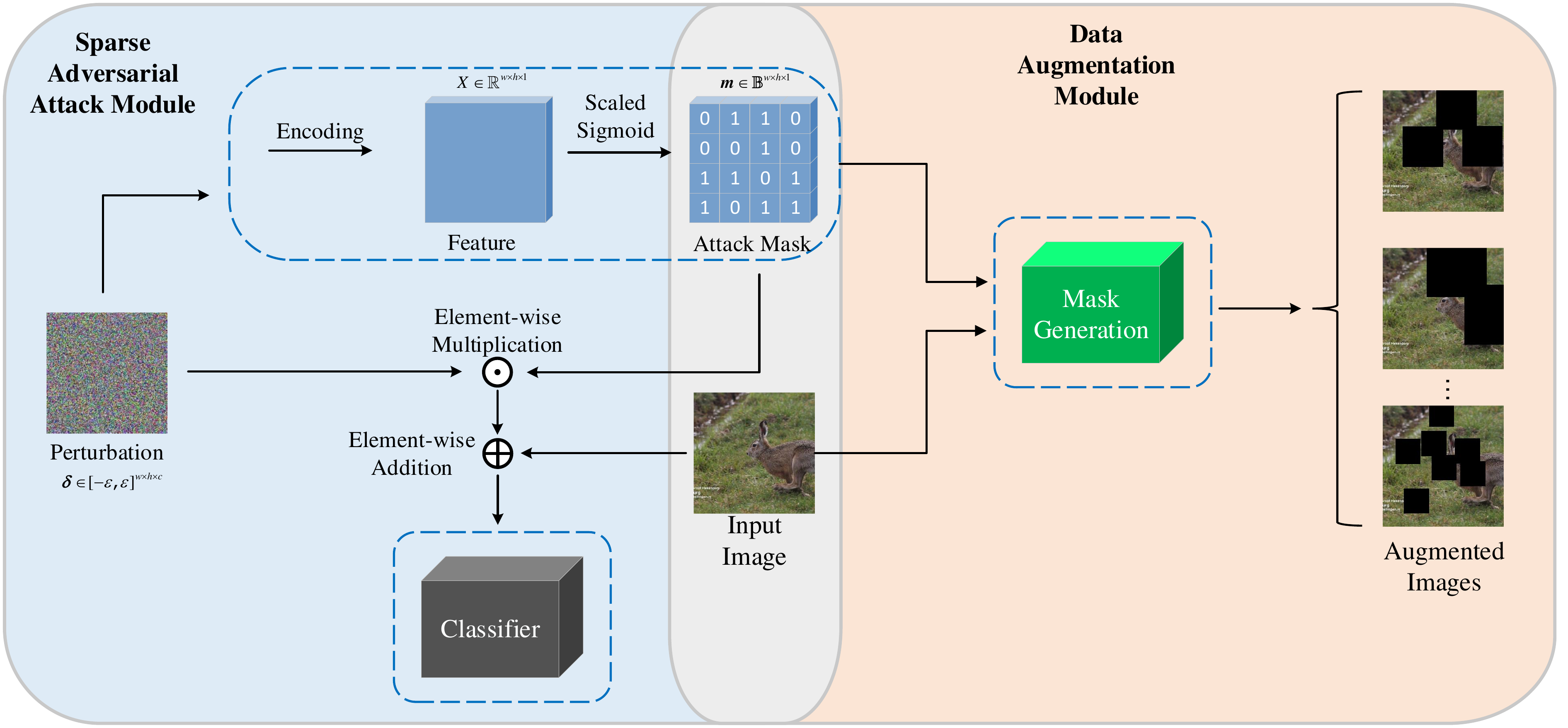}
	\caption{The framework of proposed AdvMask, which consists of two modules: sparse adversarial attack module and data augmentation module.
		Given a perturbation, we encode it through a trainable neural network to generate a feature map. Then, we use the scaled sigmoid function to generate an approximately binary attack mask.  
		The attack mask is used to guide the generation of augmentation mask for augmented images. 
		After mask generation, the shapes of generated augmented masks are very diverse. Here we show a few augmented images.}
	\label{framework}
\end{figure*}

Generally speaking, AdvMask consists of two modules: sparse adversarial attack module and data augmentation module, as illustrated in Fig.~\ref{framework}.
Using the learned sparse adversarial attack model, we can customize the augmented training samples for each image in the process of model training. 
In this way, the augmented data can simulate various situations where some key information is lost in the image, which can encourage the model to pay more attention to less sensitive areas with important features.

AdvMask can reduce the influence of over-fitting, enhance the generalization ability of classifier, and thus improve the classification accuracy.
To demonstrate the effectiveness of AdvMask, we also conducted experiments on CIFAR10/100~\cite{cifar} and Tiny-ImageNet~\cite{tiny} for coarse-grained classification, and Oxford Flower Dataset~\cite{flower} for fine-grained classification and effect visualization.  

To summarize, the contributions of this work are featured in the following aspects:
\begin{itemize}
	\item A novel data augmentation method is proposed, which innovatively uses sparse adversarial attack methods for data augmentation. 
	AdvMask first finds key information in the images through sparse adversarial attack module, and then generates customized augmented data for every image.
	\item We design an end-to-end sparse adversarial attack method, which uses the attack to guide the automatic selection of pixels without hand-crafted rules.
	Compared with other sparse adversarial attack methods, it is competitive in speed and attack success rate.
	\item We develop a data augmentation module based on occlusion suppression in AdvMask.
	By suppressing the occlusions on key information, classification models can focus more on less sensitive areas with important features.
	\item We further conduct image classification experiments to compare and discuss the proposed method with other data augmentation techniques, and analyze the parameters of our proposed method in more detail by ablation research. 
	Experimental results demonstrate that the proposed data augmentation method achieves superior performance on several datasets, and improves the performance of various popular image classification models.
	\item We are the first to propose the method of sparse adversarial attack in the community of data augmentation.
	We prove that key points from sparse adversarial attack module are important for image classification, and are very useful for designing data augmentation methods.
	%	Sparse adversarial attack is a critical research field which fools the classification models with perturbations of several key points .
\end{itemize}
The reminder of the paper is organized as follows: In Section~\ref{related-work}, we review related works about sparse adversarial attack and data augmentation methods. 
Section~\ref{advmask} elaborates on the proposed data augmentation method based on information deletion for image classification, including sparse adversarial attack module and data augmentation module. 
Section~\ref{experiment} presents the experimental results of the comparisons and evaluations.
Ablation experiment results and performance analysis are reported in Sectoin~\ref{ablation-study}.
Finally, the conclusion is drawn in Section~\ref{conclusion}. 

\section{Related Work}\label{related-work}
\subsection{Sparse Adversarial Attack}\label{related-saa}
The main difficulty of sparse adversarial attack is how to determine the pixels to be perturbed. 
Most of the existing methods can be described by a two-stage pipeline.
First, these methods artificially define a measure of pixel importance, such as gradients.
Then according to the importance of each pixel, the iterative strategy is applied, which selects the most important pixels to attack until the attack is successful.
For example, based on the saliency map, JSMA~\cite{jsma} uses a heuristic strategy to iteratively select the pixels to be perturbed in each iteration.
C\&W-$l_0$~\cite{cw} first uses the attack under the constraint of $l_2$-norm, and then fixes several least important pixels according to the perturbation magnitudes and gradients. 
PGD-$l_0+l_{\infty}$~\cite{pgdl} projects the perturbations generated by PGD~\cite{pgd} onto the $l_0$-ball to achieve the $l_0$ version of PGD. 
The specific projection method is to fix some pixels that do not need to be perturbed according to the perturbation magnitudes and projection loss. 
SparseFool~\cite{sparsefool} converts the problem into an $l_1$-norm constrained problem, and selects some pixels to perturb in each iteration according to the geometric relationship. 
Based on the gradient and the distortion map,  GreedyFool~\cite{greedyfool} selects some pixels to be added to the modifiable pixel set in each iteration, and then uses the greedy method to drop as many less important pixels as possible to obtain better sparsity.
However, all these methods are trying to find a better way to evaluate the importance of pixels, and use greedy methods to remove as many unimportant pixels as possible to achieve better sparsity.
We think that combining these two steps is a better choice.

There are also some methods that go beyond the two-stage pipeline. One-Pixel attack uses the differential evolution algorithm to explore the extreme cases where only one pixel is perturbed. 
However, this method has a low attack success rate.
SAPF~\cite{sapf} is similar to our adversarial attack method. This method factorizes the perturbation into the product of the perturbation magnitude and the binary mask, and uses $l_p$-box ADMM~\cite{admm} to optimize them jointly. Our method also uses the binary mask, but we generate the mask through a trainable neural network. Experimental results show that the adversarial attack points generated by our method have better sparsity.
\subsection{Data Augmentation Methods}
Deep neural network often suffers from severe over-fitting, which weakens its generalization ability.
Data augmentation is an effective method to solve the  problem of over-fitting, and it has long been used in practice.
The core idea of data augmentation is to generate more data on the basis of current training set.
The baseline method in this work is basic image manipulation, including cropping, translation, and rotation, which are widely used in image classification tasks.
Recently, some methods based on information deletion aim at simulating the occlusion in image classification tasks. 
Random Erasing (RE)~\cite{randomerasing} reduces the risk of over-fitting and makes the model robust to occlusion by randomly selecting a rectangular region in an image and erasing its pixels with random values.
Similarly, Hide-and-Seek (HaS)~\cite{has} randomly hides patches in a training image, which can  improve the object localization accuracy and enhance the generalization ability of CNN models.
Cutout~\cite{cutout} is a simple regularization technique of randomly masking out square regions of input images and can improve the robustness and overall performance of CNN models.
More recently, GridMask~\cite{gridmask} uses structured dropping regions in the input images, which emphasizes the importance of the balance between information deletion and reservation, and greatly improves the performance of CNN models on several CV tasks.
Similar to GridMask, FenceMask~\cite{fencemask} overcomes the difficulty of small object augmentation, and improves the performance over baselines by enhancing the sparsity and regularity of the occlusion block.
For person re-identification tasks, \cite{adv_data}  uses a sliding occlusion mask with specific horizontal and vertical strides to search for discriminative regions which are determined by some network visualization techniques.
In contrast to~\cite{adv_data}, the occluded regions of the augmented samples in our work are based on the adversarial attack points precalculated by our end-to-end sparse adversarial attack module. 

\section{AdvMask}\label{advmask}
As illustrated in the Fig.~\ref{framework}, AdvMask consists of two modules: sparse adversarial attack and data augmentation.
First, we distinguish two types of masks in AdvMask: attack mask and augmentation mask.
The attack mask is the output of the sparse adversarial attack module, and it indicates the position of the adversarial attack points. 
The augmentation mask in the data augmentation module is generated according to the attack mask, and it indicates the occluded regions of augmented data.

Firstly, AdvMask takes as input a whole image and a random adversarial perturbation. 
We then encode the perturbation by a trainable neural network to generate a binary attack mask of the same size.
From the binary attack mask, we obtain the points of interest (POI) for the data augmentation module.
Finally, AdvMask performs data augmentation based on the POI from the binary mask.
\subsection{Sparse Adversarial Attack}\label{binar}
Sparse adversarial attacks can find the most sensitive pixels in the images, because applying invisible perturbations to these pixels will lead to wrong classification.
Let $f:[0,1]^{w \times h \times c} \rightarrow \mathbb{R}^{K}$ be the classification model, where $w$, $h$ and $c$ denote the width, height, and the number of channels of the images, respectively. 
$K$ denotes the total number of classes.
The adversarial attack is generally formulated as:
\begin{equation}\label{problem1}
	\begin{array}{ll} 
		\min\limits_{\boldsymbol{\delta}} & \|\boldsymbol{\delta}\|_0\\
		\text{ s.t. } & \mathop{\arg\max}\limits_{r=1,...,K} f_r(\boldsymbol{x}+\boldsymbol{\delta}) \neq y_{true}
	\end{array}
\end{equation}
where $\boldsymbol{\delta} \in \mathbb{R}^{w \times h \times c}$ is the adversarial perturbation,  $\|\cdot \|_0$ denotes the $l_0$-norm that is the number of non-zero elements, and $f_r(\cdot)$ denotes the probability that the model classifies the input as class $r$. $\boldsymbol{x}$ and $y_{true}$ denote the input image and its true label.

However, problem \eqref{problem1} does not limit the maximum perturbation magnitude.
When the magnitude is large, the attack can be completed by perturbing only a few pixels~\cite{onepixel}. 
Too few perturbed pixels are not helpful for mask generation in the data augmentation module.
The sparsity of adversarial attack points is beneficial to data augmentation because these points are more likely to cover both foreground and background.
Therefore, it is necessary to constrain the magnitudes of sparse adversarial perturbations.
Our work uses $l_0$-norm as a distance function to learn sparse adversarial attack.
We formulate the sparse adversarial attack as:
\begin{equation}\label{problem2}
	\begin{array}{ll}
		\min _{\boldsymbol{\delta}} & \|\boldsymbol{\delta}\|_{0} \\
		\text { s.t. } 	& \boldsymbol{\delta} \in[-\epsilon, \epsilon]^{w \times h \times c}\\
		& \mathop{\arg\max}\limits_{r=1,...,K} f(\boldsymbol{x}+\boldsymbol{\delta}) \neq  y_{true}
	\end{array}
\end{equation}
 where $\epsilon$ denotes the maximum perturbation magnitude allowed, which is the $l_\infty$-norm of the perturbation.

 However, because the $l_0$-norm is not differentiable, the difficulty to optimize the problem makes sparse adversarial attack a NP-hard problem~\cite{greedyfool}. 
 As mentioned in Section~\ref{related-saa}, the existing methods use artificially defined importance indicators and greedy strategies, which may not be able to obtain the optimal results.
 Inspired by the research of neural network pruning, pruning neural network can be regarded as an optimization problem of $l_0$-norm which indicates the sparsity of convolution kernels.
 Early neural network pruning methods also use artificially defined importance index of convolution kernels~\cite{thinet, nnpruning, nnpruning2} .
 Recently, AutoPruner, an automatic pruning method on neural network, gets rid of the need to manually define the importance of convolution kernels, and automatically selects the convolution kernels that need to be kept in the process of fine-tuning the models. 
 AutoPruner adds a branch in the usual network training process, which outputs a mask to automatically select filters. 
 Based on the intrinsic similarity between network pruning and sparse adversarial attacks, we add a branch that includes encoding and binarization in the usual attack process, which also outputs a mask to automatically select pixels.	
The branch takes the adversarial perturbation as its input and generates an approximate binary mask with the same size.
According to the sparsity and attack effect, the adversarial perturbation and the encoder are jointly optimized. 
By gradually forcing the scaled sigmoid function to output binary values, the perturbation of some pixels finally becomes 0, thus ensuring the sparsity. 
\paragraph{Encoding}
Let $ \mathcal{H}: \mathbb{R}^{w \times h \times c} \rightarrow \mathbb{R}^{w \times h \times 1}$ denote the neural network, which is used to encode the perturbation.
$\mathcal{H}(\boldsymbol{\delta}) \in  \mathbb{R}^{w \times h \times 1}$ represents the tensor obtained by encoding the perturbation $\boldsymbol{\delta}$ with $ \mathcal{H}$.

When the image size is small, we directly use a fully-connected layer as the encoder, and its weights are denoted as $ \mathcal{W} \in  \mathbb{R}^{(whc)\times(wh)}$.
However, large image size causes too many weights to be trained. 
Considering that dimensions of the input and output must be the same, we use the classical image segmentation network U-net~\cite{unet} as the encoder.
\begin{figure}
	\subfloat{
		\begin{minipage}{0.18\linewidth}
			\centering
			%			\setlength{\abovecaptionskip}{-0.01cm} 
			% 			\caption{Input}
			% 			\text{Input}
			\includegraphics[width=1 \textwidth]{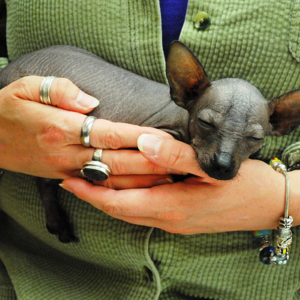}
		\end{minipage}
		\begin{minipage}{0.18\linewidth}
			\centering
			\includegraphics[width=1 \textwidth]{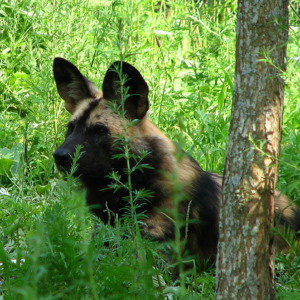}
		\end{minipage}
		\begin{minipage}{0.18\linewidth}
			\centering
			\includegraphics[width=1 \textwidth]{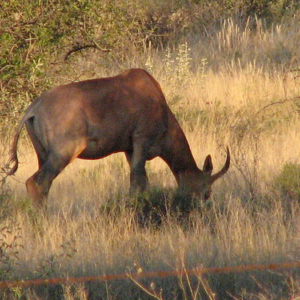}
		\end{minipage}
		\begin{minipage}{0.18\linewidth}
			\centering
			\includegraphics[width=1 \textwidth]{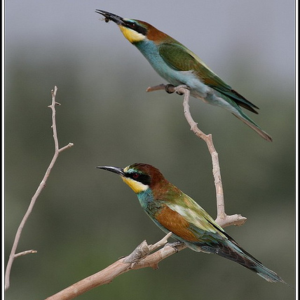}
		\end{minipage}
		\begin{minipage}{0.18\linewidth}
			\centering
			\includegraphics[width=1 \textwidth]{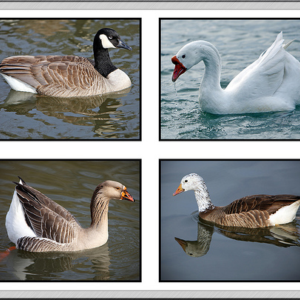}
		\end{minipage}
	}
	\newline
 
	\subfloat{
		\begin{minipage}{0.18\linewidth}
		\centering
		%			\setlength{\abovecaptionskip}{-0.01cm} 
		% 			\caption{Input}
		% 			\text{Input}
		\includegraphics[width=1 \textwidth]{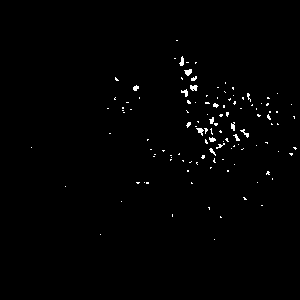}
		\end{minipage}
		\begin{minipage}{0.18\linewidth}
			\centering
			\includegraphics[width=1 \textwidth]{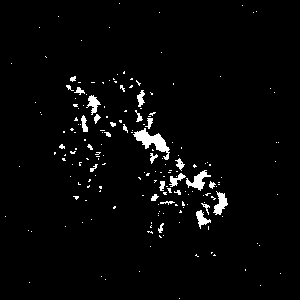}
		\end{minipage}
		\begin{minipage}{0.18\linewidth}
			\centering
			\includegraphics[width=1 \textwidth]{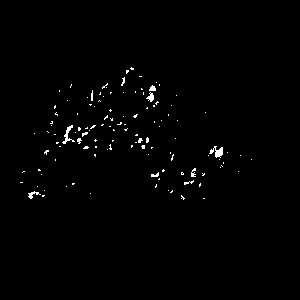}
		\end{minipage}
		\begin{minipage}{0.18\linewidth}
			\centering
			\includegraphics[width=1 \textwidth]{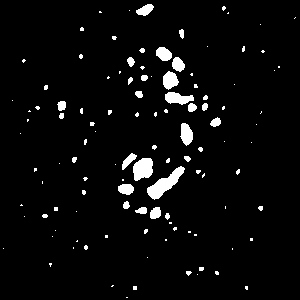}
		\end{minipage}
		\begin{minipage}{0.18\linewidth}
			\centering
			\includegraphics[width=1 \textwidth]{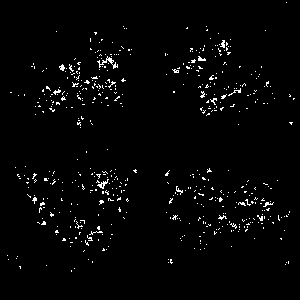}
		\end{minipage}
			}
	\caption{Illustration of attack mask generated by our sparse adversarial attack module.
	The first row: original images from ImageNet dataset~\cite{imagenet}. The second row: attack masks for images in the first row.}
	\label{attackmask}
	\end{figure}
\paragraph{Binarization}
The elements of the encoded tensor are real numbers, but the elements of the attack mask are binary, in which 1 denotes the key points and 0 denotes otherwise.
Therefore, in order to maintain continuity and differentiability, we use the scaled sigmoid function to generate an approximate binary mask:
\begin{equation}\label{sigmoid}
	\boldsymbol{m}=\operatorname{sigmoid}(\alpha \cdot \mathcal{H}(\boldsymbol{\delta}))
\end{equation}
where $	\boldsymbol{m} \in [0,1]^{w \times h \times 1}$ denotes the approximate binary mask and $\alpha$ is the scaling factor that controls the degree of binarization.
When $\alpha$ is too small, \eqref{sigmoid} is not enough to binarize the elements of mask. 
When $\alpha$ is too large, the selection of pixels has been determined before the training, which degenerates our method into randomly selecting pixels.
Therefore, we gradually increase $\alpha$ in the training process to ensure that the elements of mask can converge to binary values, and prevent the method from degenerating into random pixels selection.
\paragraph{Loss Function}
Our goal is to find the most sensitive pixels in the image while these points are sparse.
Therefore, the design of loss function is to fool the target model with the pruned perturbations as much as possible, and the key pixels should be as sparse as possible.
First we use adversarial attack, such as PGD and MI-FGSM, to calculate the loss of cross entropy with successful attack,  $L_{init}$.
We formulate our loss function for sparsity as follows:
\begin{equation}\label{eq4}
	\begin{split}
		\mathcal{L}=\max(\mathcal{L}_{\mathrm{classify}}, \eta \cdot \mathcal{L}_{\mathrm{classify}})+\lambda \frac{\|\boldsymbol{m}\|_{1}}{N},\\
		\mathcal{L}_{\mathrm{classify}} = - \frac{\mathcal{L}_{\boldsymbol{CE}}(f(\boldsymbol{x+ \delta \odot \boldsymbol{m}}),y_{true})}{L_{init}}+1
	\end{split}
\end{equation}
where $\odot$ denotes the element-wise multiplication, $\mathcal{L}_{\mathrm{classify}}$ is the adversarial classification loss, usually the cross-entropy loss.
Since the elements of $\boldsymbol{m}$ are approximately binary, the second term approximately represents the sparsity of the mask, where $N=w \times h \times 1$ represents the size of the mask. $\lambda$ is a dynamic parameter to balance these two terms, which is calculated by the following formula:
\begin{equation}
	\lambda=C+\frac{\gamma}{N} \sum_{i=1}^{N} \mathbb{I}\left(m_{i}>0.5\right)
\end{equation}
where $C>0$, a hyperparameter, is the minimum value of $\lambda$, $m_i$ is the $i$-th element of the mask, and $\gamma>0$ is a hyperparameter. 
If the current mask is not sparse enough, our method would pay more attention to improving the sparsity of the mask, otherwise our method would focus on the adversarial attack.

\paragraph{Update Perturbation and Encoder}
Since attack mask is to find sparse POI in the image, we do not need to consider its own sparsity. Therefore, instead of updating the $\boldsymbol{\delta}$ with greedy strategies, we can use adversarial attack methods, such as PGD and MI-FGSM, to quickly update $\boldsymbol{\delta}$.  
Moreover, considering the constrain of $\boldsymbol{\delta}$ in problem \eqref{problem2}, we can combine PGD and MI-FGSM~\cite{MI-FGSM} under the $l_\infty$-norm constrain.
In this way, we can constrain both the $l_0$-norm and $l_\infty$-nrom of the perturbation. Specifically, the update formula of $\boldsymbol{\delta}$ is as follows:
\begin{equation}
	\begin{aligned}
		&\boldsymbol{g}_{t+1}=\mu \cdot \boldsymbol{g}_{t}+\frac{\nabla_{\boldsymbol{\delta}} \mathcal{L}}{\left\|\nabla_{\boldsymbol{\delta}} \mathcal{L}\right\|_{1}} \\
		&\boldsymbol{\delta}_{t+1}=\operatorname{Clip}_{\boldsymbol{\epsilon}}\left\{\boldsymbol{\delta}_{t}-\beta \cdot \operatorname{sign}\left(\boldsymbol{g}_{t+1}\right)\right\}
	\end{aligned}
\end{equation}
where $\mathcal{L}$ is the loss defined by~\eqref{eq4}, $\mu$ denotes  the momentum decay factor, $\beta$ represents the update step, and $\operatorname{Clip}_{\boldsymbol{\epsilon}}\left\{ \cdot \right\}$ is used to project adversarial perturbation into the $l_\infty$-ball of radius $\epsilon$.

To summarize, the perturbation is trained simultaneously with the encoder. The completion of training means the completion of the attack mask generation. Fig.~\ref{attackmask} illustrates some examples of attack masks.

\subsection{Data Augmentation}\label{data_aug}
Excessively deleting regions in the image may lead to complete object removal and context information loss, while retaining too many regions may lead to information reduction~\cite{gridmask}.
Therefore, the key of data augmentation based on information deletion is to balance between deletion and reservation of continuous areas. 
Our proposed information removal method, AdvMask, utilizes structured dropping regions.
As is illustrated in Fig.~\ref{framework}, the attack mask and input image are used for data augmentation.
The areas removed by our method are neither continuous areas nor random areas. 
The core idea of our data augmentation module is to remove the square areas centered on the adversarial attack points from the attack mask, which is indicated by augmentation mask.
In augmentation mask $M$, if $M_{i,j}=1$, the pixel at position $(i,j)$ is unchanged; otherwise, we remove it. 

%On the contrary, on the basis of sparse adversarial attack, AdvMask removes regions centered on the sparse adversarial attack pixels, as shown in Fig.~\ref{framework}. 
%The adversarial attack module generates attack masks, and the data augmentation module removes regions with augmentation mask according to the POI.
We express our setting as:
\begin{equation}
	\mathbf{x'}=\mathbf{x}  \odot M
\end{equation}
where $\mathbf{x} \in  \mathbb{R}^{w \times h \times c}$ denotes the input image, $M \in [0,1]^{w \times h}$ is the binary augmentation mask indicating the pixels to be removed, and $\odot$ means the element-wise multiplication.

The augmentation mask $M$ is determined by five parameters:$(l_{min}, l_{max}, p_{min}, p_{max}, o_{max})$.
\paragraph{Square Length $l$} \label{para:l}
From sparse adversarial attack, we can get the position of POI of each image.  
So we can customize the augmentation masks for each image.
For each perturbation point in the attack mask, we design a square $S$ centered on this point. The parameter $l$ is the length of $S$. 
The value of $l$ is a random sample within a range:
\begin{equation}
	l = random(l_{min}, l_{max})
\end{equation}
The parameter $l$ is set according to the image sizes.
\paragraph{Mask Ratio $p$}\label{para:p}
Since some recent works~\cite{reid, gridmask} revealed that dropping a very small region is useless for convolutional operation, parameter mask ratio $p$ is used to control the proportion of total masked pixels. 
The larger $p$ is, the more pixels are masked.
Considering the balance between information deletion and reservation, the scale range of the masked regions $p$ should be limited.
Once the parameter $l$ is determined, the area of total masked regions can be calculated by:
\begin{equation}
	sum(M)= \sum_{S}   l^2
\end{equation}
where $S$ is the number of squares in the mask.  
Then, we get:
\begin{equation}
	p = \frac{sum(M)}{h \times w}
\end{equation}
where $h$ and $w$ are the height and width of the image.
$p$ can prevent algorithm from removing too many or too few pixels by limiting the number of squares in the mask. Therefore, we set a scale range of $p$ as $[p_{min},p_{max}]$.
In this case, the dropping regions can neither be too large nor too small.
\paragraph{Overlapping Ratio $o$}\label{para:o}
Since square $S$ has different length, if two perturbation points are very close, the squares centered on these two points may easily overlap.
This situation could deteriorate sharply when the lengths of the squares are large while the perturbation points are dense, which causes the deleted areas to continuously overlap and finally concentrate on a sub-areas. 
In the worst case, no matter how large $l$ is, the entire object of interest will be deleted.
Therefore, by controlling the overlapping ratio $o$, on the one hand, we can prevent the algorithm from removing continuous areas. On the other hand, we can ensure the structural characteristics of the mask.
So for every mask,  we set a upper bound of $o$, and randomly choose $o$ as 
\begin{equation}
	o = random(0, o_{max})
\end{equation}
In this way, the overlapping area of every pair of squares in the augmentation mask is no larger than $\lceil l^2 \times o \rceil$.

The random values of these five parameters increase the diversity of the augmentation masks and ensure the structural deletion of areas.
For every image in each epoch during training, we generate different masks.
Even for the same image and its attack mask, every generated augmentation mask is different.
\begin{table*}[!ht]
	\centering
	\renewcommand\arraystretch{1.3}
		\caption{Image classification accuracy on CIFAR-10 and CIFAR-100 are summarized in this table. 
			* means results reported in the original paper. \textbf{RE}: random erasing. }
	%\resizebox{0.45\textwidth}{!}{
		\resizebox{.9\textwidth}{!}{
			\begin{tabular}{c|l|llllll}
				\toprule[1.5pt]
				Dataset & Model & 				Basic & Cutout & 			HaS &		 GridMask & RE & AdvMask (ours) \\ \midrule
				& ResNet-18 &  					95.28 *& 96.01 * & 			96.10 * &	 96.38	& 	95.69 * & \textbf{96.44}\\ 
				& ResNet-44 & 					94.10 &	 94.78 &	 		94.97 &			95.02  & 			94.87 *	 & \textbf{95.49} \\
				CIFAR-10 & ResNet-50 & 95.66& 	 95.81 &			95.60 &96.15    & 			95.82& \textbf{96.69}\\
				& WideResNet-28-10& 	95.52 &  96.92 & 			96.94 &			\textbf{97.23} 	 & 96.92 &			 97.02\\ 
				&Shake-shake-26-32& 	94.90 &   96.96 *&96.89 *&			96.91	&				 96.46 *&\textbf{97.03}\\ \hline
				
				& ResNet-18 & 				 		77.54 * &  			78.04 *&	78.19&		75.23&		75.97 *&	\textbf{78.43}\\ 
				& ResNet-44 &				 		74.80& 				74.84&		75.82&			76.07&		75.71 *&	\textbf{76.44} \\
				CIFAR-100 & ResNet-50 &  77.41&					78.62		&		78.76&			78.38&		77.79&			\textbf{78.99}\\
				& WideResNet-28-10& 		78.96&		   	79.84&			80.22&			80.40&		80.57&		\textbf{80.70}  \\ 
				&Shake-shake-26-32& 		76.65&			77.37&			76.89&			77.28&		77.30&		\textbf{79.96}\\ 
				\bottomrule[1.5pt]
			\end{tabular}}
		\label{cifar}
		%	\resizebox{0.5\textwidth}{!}{}
	\end{table*}
\section{Experiment}\label{experiment}
In this section, we report on experiments to evaluate our method and compare it with other data augmentation methods.
To prove the effectiveness of our proposed method, the experiments include four image datasets and various popular neural network models in image classification tasks. 
For the same model trained with different data augmentation methods, we use the same parameter settings, such as learning rate.
\subsection{Datasets and Evaluation Metrics }
We conduct experiments on four representative datasets for image classification tasks: CIFAR-10/100, Tiny-ImageNet, and Oxford Flower Dataset.
\begin{enumerate}
	\item \textbf{CIFAR-10/100}~\cite{cifar}: Both CIFAR-10 and CIFAR-100 datasets consist of 60000 color images of size $32 \times 32$ pixels, while the former has 10 distinct classes and the latter has 100. 
	Each dataset is divided into a training set containing 50000 images and a test set containing 10000 images.
	Experimental results on these datasets demonstrate the superiority of AdvMask for coarse-grained classification on small image sizes.
	\item \textbf{Oxford Flower Dataset}:~\cite{flower} consists of 102 flower categories.
	Each class contains between 40 and 258 images of size $224 \times 224$.
	These images have large scale, and vary greatly in pose and light.
	In addition, because the images are larger in size and  clearer in vision, we present the results of class activation maps (CAM)~\cite{cam} from models trained with different data augmentation methods.
	Experiment results on the dataset demonstrate the superiority of AdvMask for coarse-grained classification on large image sizes, and visually prove the effect of the model trained with AdvMask.
	\item \textbf{Tiny-ImageNet}:~\cite{tiny} contains 100000 images of 200 classes. Each class has 500 training images, 50 validation images, and 50 test images.
	Since test images are not labeled, we use validation set to test.
	Experimental results on the dataset prove the superiority of AdvMask for fine-grained dataset with large image sizes.
	\item \textbf{Evaluation Metric}: We use Rank-1 and mean average classification accuracy to evaluate the performance of the proposed method. 
\end{enumerate}
\subsection{Implementation Details}
Since our adversarial attack module is white-box attack, we first use a trained classification model as the attack target to obtain the attack mask for every image.
Based on the attack mask, we then generate augmentation masks in data augmentation module.

For the sparse adversarial attack module, there are some important implementation details to note.
As mentioned in Secton~\ref{binar}, we gradually increase the scaling factor $\alpha$ from $\alpha_{min}$ to $\alpha_{max}$ in the optimization process to make the mask binary. Specifically, $\alpha_{max}$ is set to 100 on CIFAR-10 and CIFAR-100, and 5 on Tiny-ImageNet and Oxford Flower Dataset. The change of $\alpha_{min}$ has little effect on the results, so we just set $\alpha_{min}$ as 0.1 all the time.

For the data augmentation module, the datasets is normalized using per-channel mean and standard deviation, and our algorithm is applied after the image normalization operation. 

Inspired by the easy-to-hard learning strategy~\cite{easy2hard, easy2hard2},  we propose an incremental generative strategy.
Specifically, instead of applying AdvMask to every image in each epoch, the number of augmented occluded samples gradually increases during training, which will make the network more robust to occlusion by learning more and more occlusion samples.
The number of augmented samples grows uniformly with epoch until it reaches a constant upper bound. 
In practice, we set the upper bound to 80\% of the total sample size.
\subsection{Results on CIFAR-10 and CIFAR-100}
 We use ResNet~\cite{resnet} architecture with different size models, including ResNet-18, ResNet-44, and ResNet-50.
 In addition, we use WideResNet-28-10~\cite{wrn} and ShakeShake-26-32~\cite{shake-shake}. 
 The hyperparameters we use are the same as those reported in the paper~\cite{cutout, gridmask}.
 For the basic augmentation, we pad the image to $36 \times 36$ and randomly crop it into size $32 \times 32$. 
 The images are  normalized using per-channel mean and standard deviation. AdvMask is applied after the normalization.
 For the basic augmentation, we pad the image to $36 \times 36$ and randomly crop it into size $32 \times 32$, then  horizontally flipped with the probability of 0.5.
 The parameters for AdvMask is correlated to the complicity of datasets.
 Generally, the more complex the dataset is, the larger masked regions is preferred. Therefore, we train with CIFAR-10 using the range of square length $l$ of $[2,15]$, the range of mask ratio  $p$ of $[0.2,0.4]$, and overlapping ratio $o$ of 0.1.
 For CIFAR-100, the range of square length $l$ of $[5,20]$, the range of mask ratio  $p$ of $[0.06,0.5]$, and overlapping ratio $o$ of 0.2.
  
 We conduct the experiments several times and report the best classification accuracy of different models on CIFAR-10 and CIFAR-100, as summarized in Table~\ref{cifar}. 
 Notably, AdvMask can improve the performance of ResNet18, ResNet44, ResNet50, WideResNet-28-10 and ShakeShake-26-32 on baseline by 1.16\%, 1.39\%, 1.03\%, 1.50\%, and 2.12\%, respectively.
 Since the complexity of CIFAR-10 is not high, the accuracy of different methods is not much different.
 
 Especially for ResNet44, WideResNet-28-10, and ShakeShake-26-32, we have improved the classification accuracy on CIFAR-100 by 1.64\%, 1.74\% and 3.31\%, respectively. 
 AdvMask achieves best results on these 5 classification models, which shows the superior to previous data augmentation methods.
  Meanwhile, for average classification accuracy,  AdvMask is the state-of-the-art method among different models.
 \subsection{Results on Oxford Flower Classification Dataset}
\newcommand{\len}{1.}
 \newcommand{\Len}{0.13}
 \newcommand{\vlen}{-2.5}
 \begin{figure*}[ht]
 	\centering
	\begin{flushleft}
		\footnotesize ~~~~~~~~ \qquad\qquad Input\qquad\qquad\quad ~~~Baseline\qquad\qquad ~~~~~Cutout \qquad\qquad ~~~Gridmask~\qquad\qquad \quad ~~~RE \qquad\qquad\qquad ~~Ours
	\end{flushleft}

 	\subfloat{
 		%				 				\rotatebox[origin=c]{90}{\scriptsize{Figure}}
 		
 		\begin{minipage}{\Len\linewidth}
 			\centering
 			\setlength{\abovecaptionskip}{-0.01cm} 
% 			\caption{Input}
% 			\text{Input}
%			Input 
 			\includegraphics[width=\len \linewidth]{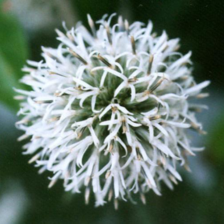}
 		\end{minipage}
 		%\qquad
 		\begin{minipage}{\Len\linewidth}
 			\centering
 			\setlength{\abovecaptionskip}{-0.01cm} 
% 			\caption{Baseline}
%			\text{Baseline}
%			Baseline
 			\includegraphics[width=\len \linewidth]{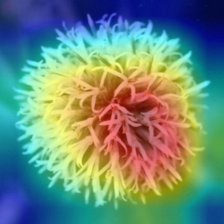}
 		\end{minipage}
 		\begin{minipage}{\Len\linewidth}
 			\centering
 			\setlength{\abovecaptionskip}{-0.01cm} 
% 			\caption*{Cutout}
 			\includegraphics[width=\len \linewidth]{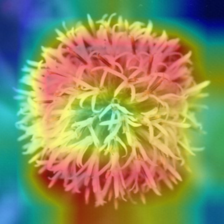}
 		\end{minipage}
 		\begin{minipage}{\Len\linewidth}
 			\centering
 			\setlength{\abovecaptionskip}{-0.01cm} 
% 			\caption*{GridMask}
 			\includegraphics[width=\len \linewidth]{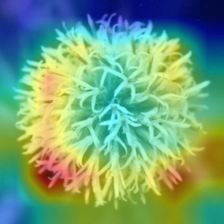}
 		\end{minipage}
 		\begin{minipage}{\Len\linewidth}
 			\centering
 			\setlength{\abovecaptionskip}{-0.01cm} 
% 			\caption*{RE}
 			\includegraphics[width=\len \linewidth]{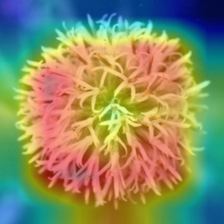}
 		\end{minipage}
 		\begin{minipage}{\Len\linewidth}
 			\centering
 			\setlength{\abovecaptionskip}{-0.01cm} 
% 			\caption*{Ours}
 			\includegraphics[width=\len \linewidth]{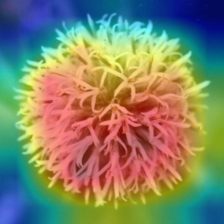}
 		\end{minipage}
 	}
 	\vspace{\vlen mm }
 	\newline
 	\subfloat{
 		%					\rotatebox[origin=c]{90}{\scriptsize{Figure}}
 		\begin{minipage}{\Len \linewidth}
 			\centering
 			\includegraphics[width=\len \linewidth]{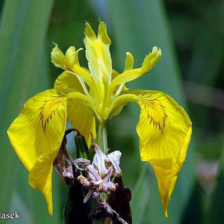}
 		\end{minipage}
 		
 		%\qquad
 		\begin{minipage}{\Len\linewidth}
 			\centering
 			\includegraphics[width=\len \linewidth]{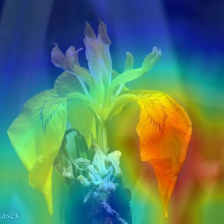}
 		\end{minipage}
 		\begin{minipage}{\Len\linewidth}
 			\centering
 			\includegraphics[width=\len\linewidth]{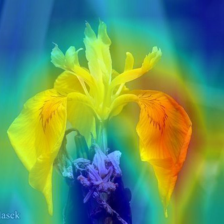}
 		\end{minipage}
 		\begin{minipage}{\Len \linewidth}
 			\centering
 			\includegraphics[width=\len\linewidth]{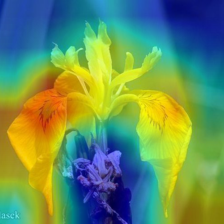}
 		\end{minipage}
 		\begin{minipage}{\Len\linewidth}
 			\centering
 			\includegraphics[width=\len\linewidth]{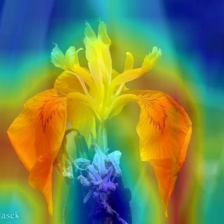}
 		\end{minipage}
 		\begin{minipage}{\Len \linewidth}
 			\centering
 			\includegraphics[width=\len\linewidth]{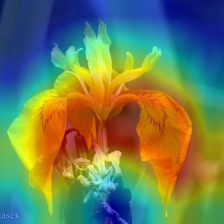}
 		\end{minipage}
 	}
 	\vspace{\vlen mm }
 	\newline
 	\subfloat{
 		%					\rotatebox[origin=c]{90}{\scriptsize{Figure}}
 		\begin{minipage}{\Len\linewidth}
 			\centering
 			\includegraphics[width=\len\linewidth]{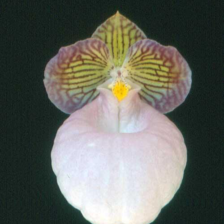}
 		\end{minipage}
 		
 		%\qquad
 		\begin{minipage}{\Len\linewidth}
 			\centering
 			\includegraphics[width=\len\linewidth]{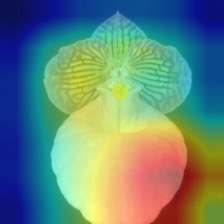}
 		\end{minipage}
 		\begin{minipage}{\Len\linewidth}
 			\centering
 			\includegraphics[width=\len \linewidth]{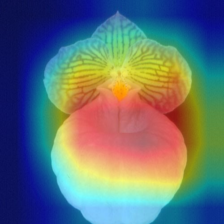}
 		\end{minipage}
 		\begin{minipage}{\Len \linewidth}
 			\centering
 			\includegraphics[width=\len\linewidth]{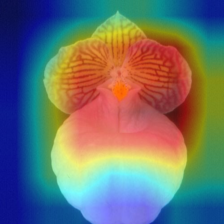}
 		\end{minipage}
 		\begin{minipage}{\Len\linewidth}
 			\centering
 			\includegraphics[width=\len\linewidth]{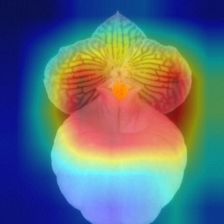}
 		\end{minipage}
 		\begin{minipage}{\Len\linewidth}
 			\centering
 			\includegraphics[width=\len\linewidth]{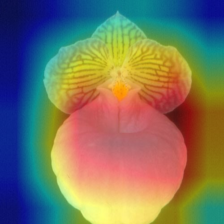}
 		\end{minipage}
 	}
 	\vspace{\vlen mm }
 	\newline
 	\subfloat{
 		%					\rotatebox[origin=c]{90}{\scriptsize{Figure}}
 		\begin{minipage}{\Len\linewidth}
 			\centering
 			\includegraphics[width=\len\linewidth]{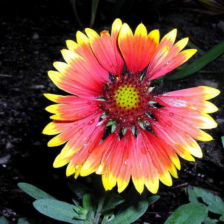}
 		\end{minipage}
 		%\qquad
 		\begin{minipage}{\Len\linewidth}
 			\centering
 			\includegraphics[width=\len\linewidth]{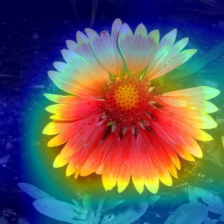}
 			
 		\end{minipage}
 		\begin{minipage}{\Len\linewidth}
 			\centering
 			\includegraphics[width=\len\linewidth]{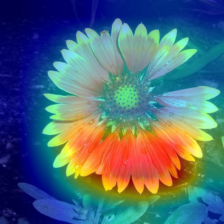}
 		\end{minipage}
 		\begin{minipage}{\Len\linewidth}
 			\centering
 			\includegraphics[width=\len\linewidth]{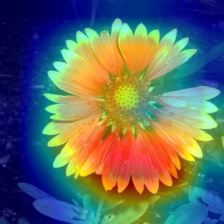}
 		\end{minipage}
 		\begin{minipage}{\Len\linewidth}
 			\centering
 			\includegraphics[width=\len\linewidth]{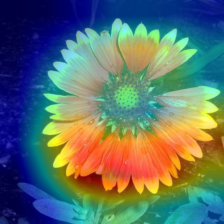}
 		\end{minipage}
 		\begin{minipage}{\Len\linewidth}
 			\centering
 			\includegraphics[width=\len \linewidth]{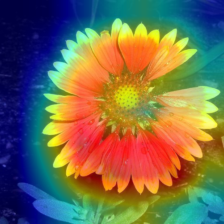}
 		\end{minipage}
 	}
 	\vspace{\vlen mm }
 	\newline
 	\subfloat{
 		%					\rotatebox[origin=c]{90}{\scriptsize{Figure}}
 		\begin{minipage}{\Len\linewidth}
 			\centering
 			\includegraphics[width=\len\linewidth]{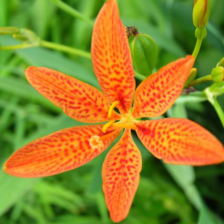}
 		\end{minipage}
 		%\qquad
 		\begin{minipage}{\Len\linewidth}
 			\centering
 			\includegraphics[width=\len\linewidth]{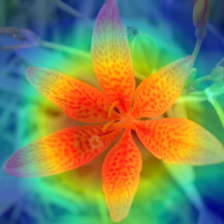}
 		\end{minipage}
 		\begin{minipage}{\Len\linewidth}
 			\centering
 			\includegraphics[width=\len\linewidth]{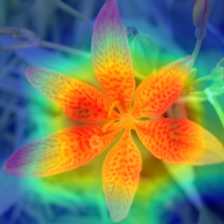}
 		\end{minipage}
 		\begin{minipage}{\Len\linewidth}
 			\centering
 			\includegraphics[width=\len\linewidth]{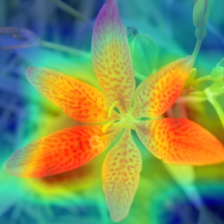}
 			
 		\end{minipage}
 		\begin{minipage}{\Len\linewidth}
 			\centering
 			\includegraphics[width=\len\linewidth]{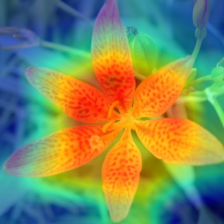}
 			
 		\end{minipage}
 		\begin{minipage}{\Len\linewidth}
 			\centering
 			\includegraphics[width=\len\linewidth]{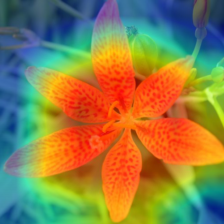}
 		\end{minipage}
 	}
 	\vspace{\vlen mm }
 	\newline 
 	\caption{Class activation mapping (CAM) for ResNet50 model trained on , with baseline augmentation, Cutout, GridMask, Randomerasing (RE) or our AdvMask. The models trained with AdvMask  incline to focus on large important regions and cover larger area of object of interest.}
 	\label{CAM}
 \end{figure*}
We randomly divide the Oxford FLower Classification Dataset into training set, validation set and test set, and use 
cross validation for experiment.
% Since the size of the original test set of Oxford FLower Classification Dataset is much larger than that of training set, which is 6129 and 1030 respectively, we use the test set for training and use the training set for testing.
 The basic augmentation includes random padding, cropping and horizontal flipping. 
 
 From Table.~\ref{flower}, in terms of accuracy, AdvMask achieves the state-of-the-art classification accuracy on this dataset.
 Specifically, compared with baseline, AdvMask improve the accuracy by 11.37\%.
%  In order to further analyze what is learned with our AdvMask-trained model, we conduct experiments with Oxford Flower Dataset.
 In addition, because the images are larger in sizes and clearer in visual effect than CIFAR-10/100, we visualize the CAM on ResNet-50 model trained with different data augmentation methods for comparison.
 It can be seen from Fig.~\ref{CAM} that AdvMask is more inclined to locate and cover relevant parts of the objects than other data augmentation methods. 
 At the same time, AdvMask can indeed achieve the best classification accuracy on ResNet-50 model, which shows that successful data augmentation help models to focus on the discriminative and salient areas in the image.
 Specifically, in the first row of Fig.~\ref{CAM}, the region of interest (ROI) of baseline method just covers part of the flower, even focuses on the background information, which is a manifestation of over-fitting problem.
On the contrary, the attention area of the models trained with AdvMask covers almost the whole flower, while the background is neglected, which proves that AdvMask improves the generalization ability of the models.
 \begin{table}[h]
 	\centering
 	 	\caption{Image classification accuracy on Oxford Flower Classification Dataset. }
 	\resizebox{.65\columnwidth}{!}{
 		\begin{tabular}{lr}
 			\toprule[1.5pt]
 			Method & Accuracy (\%) \\  \midrule
 			Basic & 80.20\\
 			Cutout & 88.53\\
 			HaS & 84.12 \\
 			RE & 88.43 \\
 			GridMask & 90.19 \\
 			AdvMask (Ours) & \textbf{91.57}\\
 			\bottomrule[1.5pt]
 	\end{tabular}}
 	\label{flower}
 \end{table}

\subsection{Results on Tiny ImageNet}
\begin{figure}[h]
	\centering
	\includegraphics[width=0.9\columnwidth]{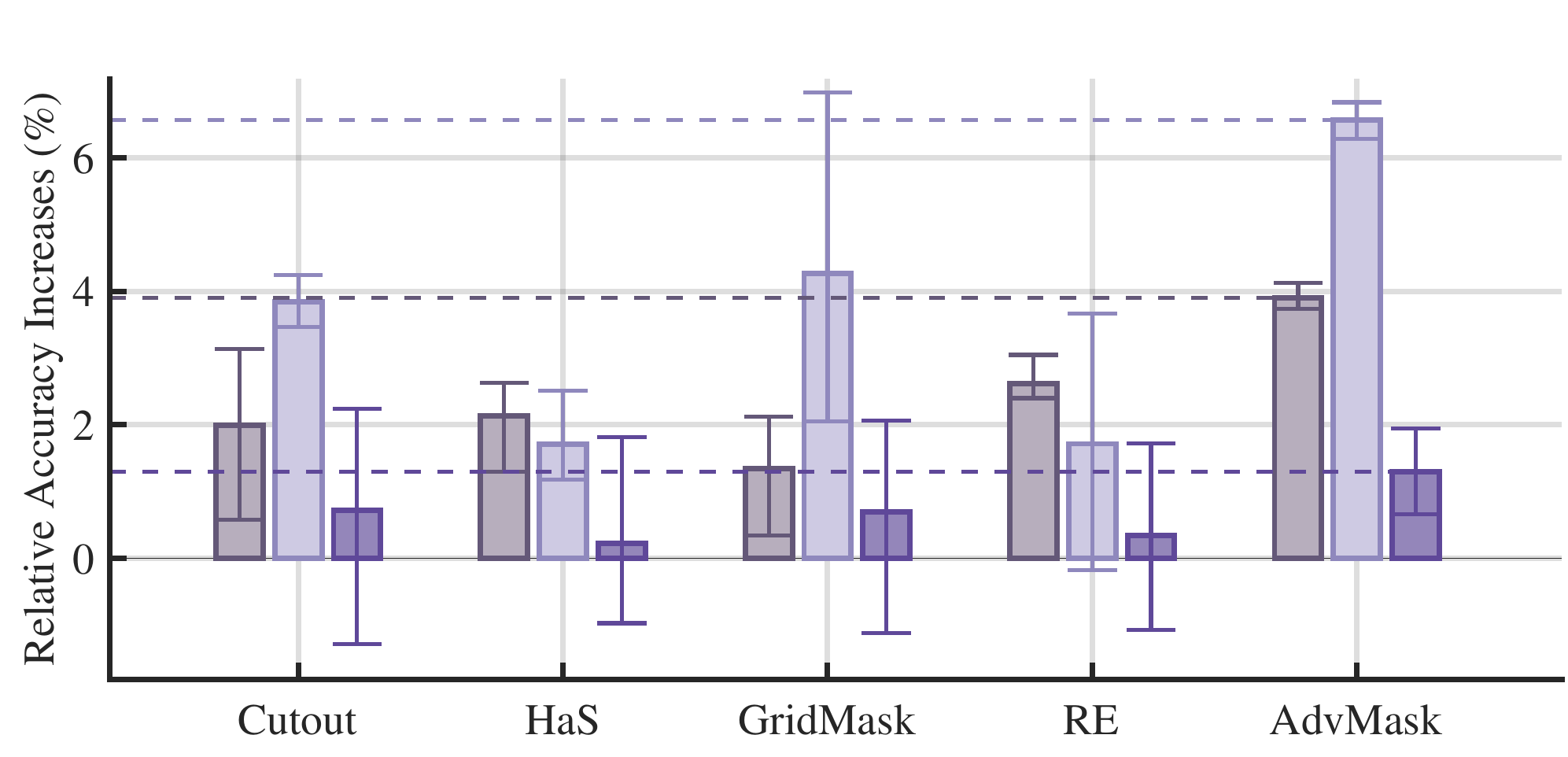}
	\caption{Relative classification accuracy compared with baseline on ResNet-18, ResNet-50, and WideResNet-50-2 for Tiny-ImageNet dataset.
	}
	\label{tiny-acc}
\end{figure}
We also conduct experiments on a larger image dataset Tiny ImageNet.
First, we resize the Tiny-ImageNet data into $64 \times 64$ and initialize all neural models with ImageNet pre-trained weight, and then fine-tune the Tiny-ImageNet dataset for some epochs. 
Therefore, we modify the network architectures to fit the image size and the output of 200 classes, including ResNet-18, ResNet-50 and Wide-ResNet 50-2~\cite{wrn}. 
The basic augmentation includes random padding, cropping and horizontal flipping. 
Images are all normalized to zero, and then AdvMask is applied.

In order to reflect the classification accuracy improvements brought by different data augmentation methods on baseline, in this experiment, we show the relative average accuracy improvements of different data augmentation methods compared with the baseline in Fig.~\ref{tiny-acc}.
The baseline accuracies of these three models are 61.38\%, 73,61\% and 81.55\%, respectively.

From Fig.~\ref{tiny-acc}, we can see that the average accuracy of all data augmentation methods is better than that of baseline. 
Among all data augmentation methods, AdvMask achieves the highest improvement in accuracy.
On ResNet-18, ResNet-50 and WideResNet-50-2, AdvMask improves the accuracy by 3.91\%, 6.57\%, and 1.30\%, respectively.
In addition, we also present the error interval of each method.  
Specifically, for these three classification models, AdvMask has the smallest error interval among all the methods, which proves that the effect of AdvMask is the most stable.
For ResNet-18 and ResNet-50, the worst-cases performance is still higher than baseline, which proves the effect of these data augmentation methods.
For WideResNet-50-2, although the worst-cases performance of other methods is worse than the average performance of baseline, the worst-cases performance of AdvMask is still better than that of baseline. Therefore, AdvMask is stable and effective.

 \section{Ablation Studies}\label{ablation-study}

 \subsection{Attack Success Rate of Sparse Adversarial Attack Module}
\begin{table}[]
 	\caption{Results of targeted sparse adversarial attack on CIFAR-10.}
 	\label{attack}
 	\centering
 	\renewcommand\arraystretch{1.5}
 	\scalebox{.9}{
 		\begin{tabular}{llrrrr}
 			\toprule[1.5pt]
 			Threshold          & Method        & ASR(\%)        & $l_0$      & $l_2$    & $l_{\infty}$  \\ \midrule
 			&JSMA & 78.9& 440.8 & 0.611 & 0.031 \\
 			& PGD-$l_0+l_\infty$      & 73.9           & 1199.7     &1.078      & 0.031          \\
 			$\epsilon=8/255$	& GreedyFool    & 100.0          & 468.2     & 0.547         & 0.031   \\
 			& C$\&$W-$l_0$ & 100.0 & 326.6 & 0.542 & 0.068\\
 			&SAPF& 100.0 & 321.8 & \textbf{0.523} & 0.085 \\ 
 			& \textbf{Ours} & \textbf{100.0} & \textbf{320.1} & 0.532 & \textbf{0.031} \\ \midrule
 			
 			&JSMA & 97.3& 247.7 & 0.896 & 0.063 \\
 			& PGD-$l_0+l_\infty$      & 72.8          & 498.0     & 1.390      & 0.063          \\
 			$\epsilon=16/255$	& GreedyFool    & 100.0          & 238.3     & 0.707         & 0.063   \\
 			& C$\&$W-$l_0$ & 100.0 & 136.7 & 0.691 & 0.118\\
 			&SAPF& 100.0 & 133.7 &  0.718 & 0.159 \\ 
 			& \textbf{Ours} & \textbf{100.0} & \textbf{131.3} & \textbf{0.689} & \textbf{0.063} \\  
 			\bottomrule[1.5pt]
 		\end{tabular}
 	}
 \end{table}
In our proposed method, the sparse adversarial attack module is used to find the key points in the images for the data augmentation module. 
Therefore, it is very important to successfully find the most important points that are most susceptible to disturbance.
We evaluate the overall performance of our  adversarial attack module through attack success rate (ASR) and $l_p$-norm ($p=0, 2, \infty$).
ASR is the proportion of misclassified adversarial samples in the total samples.
$l_0$-norm is the number of non-zero elements. The lower it is, the fewer pixels are used to apply the perturbations.
$l_2$ is the distance between the adversarial image and the original clean image.
$l_\infty$ indicates the maximum value of the perturbation magnitude, and larger it is, the greater the pixels change.
%which assesses the ability of our adversarial attack module to find the key points in the images.
%$l_p$-norm is  used to calculate the number of perturbed pixels, the distance between the original samples and adversarial samples, and the maximal perturbation magnitude.

In this subsection, we compare the effect of our proposed adversarial attack module with several state-of-the-art sparse adversarial attack methods on CIFAR-10 and ImageNet~\cite{imagenet} dataset under different parameter settings.
The sparse adversarial attack methods used include  four two-stage methods, such as JSMA, C$\&$W-$l_0$, PGD-$l_0+l_{\infty}$, GreedyFool, and one-stage method, SAPF. 
%  Different from the adversarial attack module in AdvMask using non-targeted attacks, we use targeted attacks to evaluate the ASR and compare the adversarial attack module with several state-of-the-art sparse adversarial attack methods, including four two-stage methods, such as JSMA, C$\&$W-$l_0$, PGD-$l_0+l_{\infty}$, GreedyFool, and one-stage method, SAPF. 
 The target classifier on CIFAR-10 is VGG19 model~\cite{vgg19} with the input size of $32\times32\times3$. 
 The average $l_p$ norm and ASR under two $\epsilon$ settings are shown in Table \ref{attack}, and $\epsilon$ is the maximum perturbation magnitude.
 For $\epsilon=8/255$, our proposed sparse adversarial attack module can achieve 100\% ASR and the $l_0$ is only 320.1, which means that only 10.42\% pixels are perturbed.
 For large perturbation magnitudes, our method achieves 100\% ASR and the $l_0$-norm is only 131.3 (4.27\% pixels).
 We conduct experiments of JSMA and GreedyFool, and specify the $l_\infty$-norm $= \epsilon$.
 The ASR of these two methods is not higher than ours, while the $l_0$-norm and $l_2$-norm are obviously higher than ours, which means that they have to perturb more points to successfully attack.
 Because PGD-$l_0+l_{\infty}$ needs to specify both $l_0$-norm and $l_\infty$-norm, we experiment with these two norms under the condition that $l_\infty = \epsilon$ to find the appropriate $l_0$-norm.
 However, even a larger $l_0$-norm can not achieve 100\% ASR.
 Both C$\&$W-$l_0$ and SARF can only specify the $l_0$-norm, so we conduct experiments with reference to the $l_0$-norm of our method.
 Although the $l_0$-norm of these two methods is almost the same as that of other methods, the $l_\infty$-norm of our method is much smaller than these two methods. 
 Therefore, the maximum generated perturbation magnitudes are significantly smaller.
 
 We also experiment on ImageNet, as illustrated in Table~\ref{tab:ImageNet}.
 When $\epsilon=4/255$, our method obtains 100\% ASR and the $l_0$-norm is just 12855.0, which is the lowest (4.76\% pixels). When $\epsilon=8/255$, our method can also achieve 100\% ASR, while the $l_0$-norm is the lowest (2.08\% pixels). 
 In addition, since the $l_\infty$-norm for PGD-$l_0+l_{\infty}$ and GreedyFool can be specified, we conduct the experiments under the same settings.
 We can see that the ASR of PGD-$l_0+l_{\infty}$ is low and the $l_p$-norm of these two methods is larger than ours.
   \begin{table}[h]
	\caption{Results of Targeted Sparse Adversarial Attack on ImageNet.}
	\label{tab:ImageNet}
	\centering
	\renewcommand\arraystretch{1.5}
	\scalebox{.9}{
		\begin{tabular}{llrrrr}
			\toprule[1.5pt]
			Threshold          & Method        & ASR(\%)        & $l_0$                    & $l_2$          & $l_{\infty}$  \\ \midrule
			& PGD-$l_0+l_\infty$      & 50.0           & 11989.0                    & 2.957          & 0.031          \\
			$\epsilon=4/255$	& GreedyFool    & 100.0          & 12454.5                   & 2.663          & 0.031          \\
			& \textbf{Ours} & \textbf{100.0} & \textbf{5591.5}    & \textbf{2.223} & \textbf{0.031} \\ \midrule
			& PGD-$l_0+l_\infty$      & 62.2           & 22980.9                  & 2.116          & 0.016          \\
			$\epsilon=8/255$	& GreedyFool    & 100.0          & 26107.7                 & 2.083          & 0.016          \\
			& \textbf{Ours} & \textbf{100.0} & \textbf{12855.0}  & \textbf{1.694} & \textbf{0.016} \\ 
			\bottomrule[1.5pt]
		\end{tabular}
	}
\end{table}
These results above all demonstrate the superiority and effectiveness of our proposed adversarial attack module that can indeed find sparse and most sensitive points.
%\subsection{The Effect of Adversarial Attack Points}
%To demonstrate the effect of adversarial attack points, we replace the attack points with other key points or even random points in the image, while the data augmentation module remains the same.
%First, we randomly select the points in the image as key points to apply data augmentation.
%Secondly, the data augmentation module uses the corner points as key points to generate the mask.
%Results are shown in Table ~\ref{rp_cp}. 
%Although these two point selection schemes can improve the classification accuracy on baseline, on both ResNet-18 and ResNet-50 model, our method is superior to the these two point selection schemes.
%Specifically, on ResNet-50, accuracy is 1.08\% and 1.28\% higher than random points selection and corner points selection, respectively.
%So this experiment proves the effect of adversarial attack points.
\subsection{The Effect of Generated Mask on the Targeted Points}
For AdvMask, on the one hand, the points used for mask generation are obtained by our adversarial attack module.
In order to verify the effect of adversarial attack points for data augmentation, we experiment with two point selection strategies, and then use our data augmentation module under the same experimental settings. 1) Random points (RP) selection: We randomly select points in the whole image. 2) Corner points (CP): We use corner detection to find the key points in the image.
Results are summarized in Table~\ref{rp_cp}. 
Although both of these point selection schemes can improve the classification accuracy on baseline,  our method is obviously superior to the these two point selection strategies on ResNet-18 and ResNet-50.
On Resnet-50, AdvMask is 1.08\% and 1.28\% higher in accuracy than RP and CP, respectively.
The results show the effectiveness of our adversarial attack module.

On the other hand, the data augmentation module generates various masks according to the adversarial attack points. 
In order to verify the effectiveness of data augmentation module, we conduct experiments which use attack masks directly as augmentation masks.
Therefore, the masked points are all adversarial attack points with $1 \times 1$ squares.\\
\begin{table}[]
	\caption{ Accuracy (\%) with different points selection strategies with CIFAR-10 on both Resnet-18 and Resnet-50. \textbf{RP}: random points, \textbf{CP}: corner points.}
	\label{rp_cp}
	\centering
	%\resizebox{0.45\textwidth}{!}{
		\resizebox{1.0\columnwidth}{!}{
			\begin{tabular}{l|c|c}
				\toprule[1.5pt]
				Method & ResNet-18 Acc (\%)&  ResNet-50 Acc (\%) \\
				\hline  
				\midrule
				RP & 95.66 & 95.61 \\
				CP & 95.11  & 95.41  \\
				Baseline & 95.28 & 94.12 \\ 
				\textbf{AdvMask }  & $\mathbf{96.23}$  & $\mathbf{96.69}$\\
				\bottomrule[1.5pt]
			\end{tabular}
		}
	\end{table}
\begin{table}[]
	\caption{ Test accuracy on CIFAR-10 on Resnet-18 and Resnet-50 with no mask generation data augmentation method. AAPM: adversarial attack point mask.}
	\label{aapm}
	\centering
	%\resizebox{0.45\textwidth}{!}{
		\resizebox{1.0\columnwidth}{!}{
			\begin{tabular}{l|c|c}
				\toprule[1.5pt]
				Method & ResNet-18 Acc (\%) & ResNet-50 Acc (\%) \\
				\hline 
				\midrule
				AAPM & 92.96 & 94.52 \\
				Baseline & 95.28 &   94.12 \\ 
				\textbf{AdvMask}  & $\mathbf{96.23}$ &   $\mathbf{96.69}$\\
				\bottomrule[1.5pt]
			\end{tabular}
		}
	\end{table}

From Table~\ref{aapm}, we can see that the accuracy of using attack mask is obviously lower than that of our method. 
 The accuracy is even worse than baseline on ResNet-18, because the masked region is so small and thus bring noises into the images. 
Influenced by adversarial attack, the classifier misclassifies many samples, which leads to side effect.
Our data augmentation module makes proper use of these adversarial attack points and obtains best accuracy.
\subsection{The Impact of Parameters}
\begin{figure}[h]
	\centering
	\includegraphics[width=0.95\columnwidth]{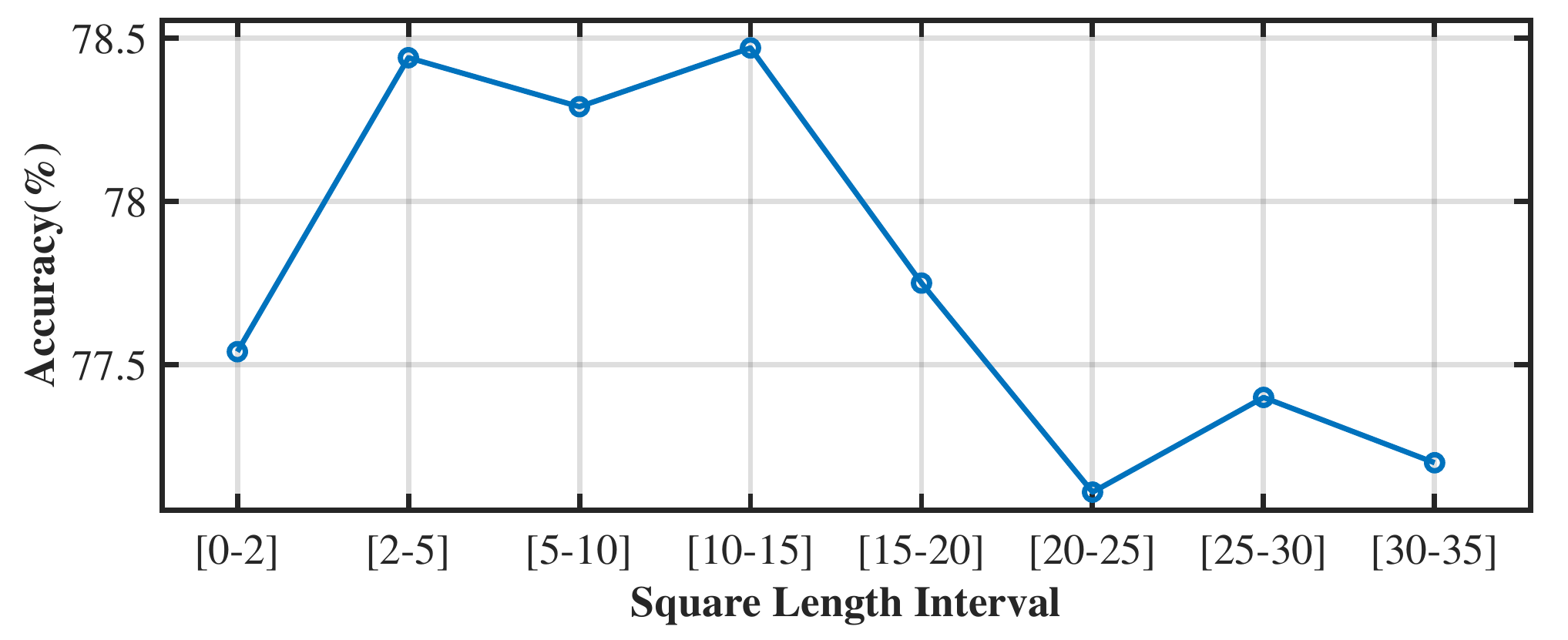}
	\caption{Classification accuracy with different settings of parameter $l$.	}
	\label{para_l}
\end{figure}

When implementing AdvMask on CNN training,  we use three parameters to control the mask generation:  square length $l$, mask ratio $p$, and overlapping ratio $\theta$, which is introduced in Section~\ref{data_aug}.
In this subsection, to demonstrate the impact of these parameters on the model performance, we conduct experiments on CIFAR-100 under varying  parameter settings. 
When evaluating one of the parameters, we fixed the other two parameters and the model architecture.
The fixed values of $l$, $p$, and $o$ are: $[2,20]$, $[0.03, 0.5]$, and $0.2$, respectively, if not specified.
\paragraph{\textbf{Parameter $\boldsymbol{l}$}}
Since the image size of CIFAR-100 is $32\times32$ and AdvMask will pad all images into larger area,  we  the maximum mask square length 35. 
Considering that  $l$ is a range interval, we set eight disjoint intervals: $[0,2]$, $[2,5]$, $[5,10]$, $[10,15]$, $[15,20]$, $[20,25]$, $[25,30]$, $[30,35]$.
Under this setting, we can analyze the effect of $l$ in each interval. Results are shown in Fig.~\ref{para_l}.
Notably, there is a trend of increasing at first and then decreasing.
If the mask square length is too small or large, the accuracy will be lower, especially when the mask is too large.
Specifically, if $l$ is lower than 2, the accuracy is far lower than that with $l$ in the range of $[2,5]$, which proves that dropping a very small region is useless for convolutional operation, and may even bring noises into the images.
When $l$ is in the interval $[2,15]$, the accuracy does not change much. 
However, if $l$  is further increased, the accuracy drops dramatically.
If $l \geq 20$, we get the lowest accuracy, even $1.36\%$ lower than the best accuracy.
$l\geq 20$ means that nearly more than 50\% of the image have been removed, so the main object in the image are likely to be removed. 

On the one hand, when the length of mask square is too small, the mask area is too small to occlude the object of interest in the image. 
On the other hand, if the mask area is too large, the whole object of interest is more likely to be totally covered, which will bring side effects to the classifier. 
For example, if $l \geq 32$, even the whole image is masked, which has a negative impact on the accuracy.
Therefore, the maximum value of $l$ should not be too great while the variety of $l$ is beneficial.
%Generally speaking, since the objects of interest have various sizes, the variety of $l$ is beneficial.
\paragraph{\textbf{Parameter $\boldsymbol{p}$}}
% $p$ is also a range interval that controls the total area of mask regions. 
Similar to the parameter $l$, we control the total mask area not to be too large or too small by adjusting $p$, so that the object can neither be completely removed nor completely kept.
We set 5 intervals, each with length 20\%, from 0 to 100\%. 

\begin{figure}[h]
	\centering
	\includegraphics[width=0.95\columnwidth]{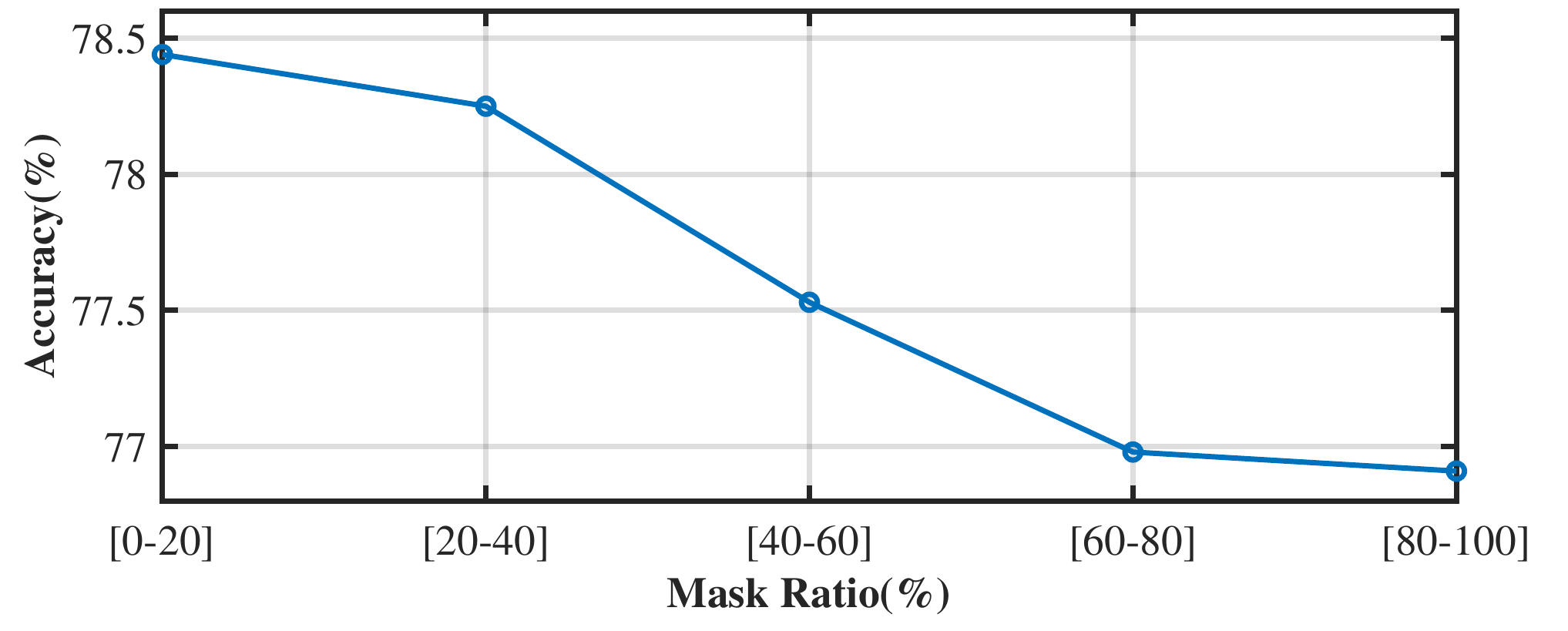}
	\caption{Classification accuracy with different settings of parameter $p$}
	\label{para_p}
\end{figure}

Results are summarized in Fig.~\ref{para_p}. 
When $p$ is relatively small, the difference in accuarcy is not significant.
However, when $p$ further increases, accuracy drops sharply.
Specifically, when $p$ is highest, we obtain the lowest accuracy. 
The classification accuracy varies by up to 1.53\%.
This is because the total masked area is so large that most of the image is masked.
As a result, the main object in the image is also removed, which will greatly damage the performance of the classifier.

\paragraph{\textbf{Parameter $\boldsymbol{o}$}} 
Because structured deletion is beneficial, we avoid deleting continuous regions in a single image.
Therefore, we experiment with different values of $o_{max}$: 0, 20\%, 40\%, 60\%, 80\%, 100\%.
$o_{max} = 0$ means that there are no overlapping area between every pair of mask squares and $o_{max} = 100\%$ means that any two mask squares can completely overlap.

\begin{figure}[h]
	\centering
	\includegraphics[width=0.95\columnwidth]{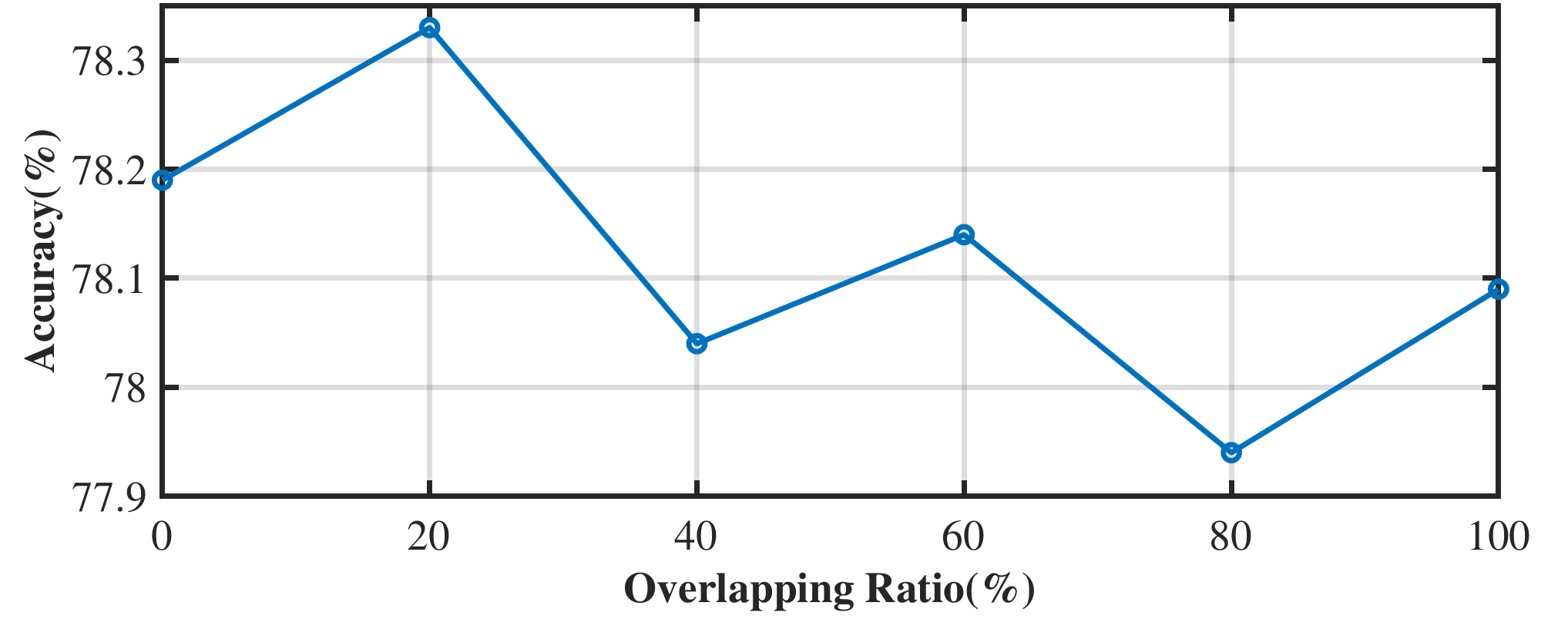}
	\caption{Classification accuracy with different settings of parameter $o$}
	\label{para_o}
\end{figure}

As illustrated in Fig.~\ref{para_o}, under the control of both $l$ and $p$, the impact of $o$ is not so significant.
%Although the change of accuracy with the increase of $o$ is not that obvious, Fig.~\ref{para_o} demonstrates the effect of $o$.
When the overlapping ratio is within the range $0-20\%$, we obtain the relatively better accuracy.
When we further increase the $o$, we get lower accuracies, with a maximum drop of 0.39\%.
From this experiment, we can see that smaller $o$ is beneficial.
This is because that with lower $o$, we can avoid deletion of continuous regions and preserve the structured information of the images. 
As a result, the main object is less likely to be fully removed nor reserved.

Experiments on these three parameters verify that any one or two parameters are sufficient to control the AdvMask. 
Therefore, we design $p$, $l$ and $o$ to control the generation of masks, so as to balance the object occlusion and information retention.
\section{Conclusion}\label{conclusion}
In this paper, we propose a data augmentation method, AdvMask, for image classification.
%Firstly, a novel sparse adversarial attack method is employed to generate key points for each image.
AdvMask first makes use of adversarial attack research to find key points in the image and then randomly occludes a structured region based on key points during each iteration. 
Neural network models trained with AdvMask can reduce the sensitivity to occlusion and effectively strengthen the generalization ability by paying more attention to less sensitive areas with important features.
The experiment results on various image datasets demonstrate the robustness and effectiveness of our method.
%From experiment results,  we can obtain reasonable improvements compared with other popular data augmentation methods.

In the future work, we will apply our data augmentation method to other CV tasks, such as, object detection and segmentation.
Finally, we hope our work provides a new research direction for the development of data augmentation research.

\section*{Acknowledgments}
This work was supported in part by the National Key R\&D Program of China under Grant 2021ZD0201300, and by the National Science Foundation of China under Grant 61876076.


\bibliographystyle{IEEEtran}
\bibliography{tip}

%\newpage

% your own custom macro containing the $\backslash${\tt{includegraphics}} command to make things
%\bf{If you include a photo:}\vspace{-33pt}
%\begin{IEEEbiography}[{\includegraphics[width=1in,height=1.25in,clip,keepaspectratio]{fig1}}]{Michael Shell}
%Use $\backslash${\tt{begin\{IEEEbiography\}}} and then for the 1st argument use $\backslash${\tt{includegraphics}} to declare and link the author photo.
%Use the author name as the 3rd argument followed by the biography text.
%\end{IEEEbiography}

%\vfill

\end{document}